\documentclass[11pt]{article}

\usepackage[final]{acl}

\usepackage{times}
\usepackage{latexsym}

\usepackage[T1]{fontenc}

\usepackage[utf8]{inputenc}

\usepackage{microtype}

\usepackage{inconsolata}

\usepackage{graphicx}
\usepackage{times}
\usepackage{soul}
\usepackage{url}
\usepackage[utf8]{inputenc}
\usepackage{subcaption}
\usepackage{CJKutf8}
\usepackage[table]{xcolor}
\usepackage{booktabs}

\title{Enhancing Multimodal Large Language Models for Ancient Chinese Character Evolution Analysis via Glyph-Driven Fine-Tuning}

\author{
Rui Song$^{1,2}$, Lida Shi$^{2,3}$, Ruihua Qi$^{2,4}$, Yingji Li$^{1}$, Hao Xu$^{1,2}$\thanks{Corresponding author.} \\
$^{1}$College of Computer Science and Technology, Jilin University, China \\
$^{2}$Key Laboratory of Ancient Chinese Script, Culture Relics and Artificial Intelligence, Jilin University, China \\
$^{3}$School of Artificial Intelligence, Jilin University, China \\
$^{4}$School of Archaeology, Jilin University, China \\
\texttt{
\{songrui,yingjili,xuhao\}@jlu.edu.cn,} \\
\texttt{
\{shild21,qirh20\}@mails.jlu.edu.cn
}
}

\begin{document}
\maketitle
\begin{abstract}
In recent years, rapid advances in Multimodal Large Language Models (MLLMs) have increasingly stimulated research on ancient Chinese scripts. As the evolution of written characters constitutes a fundamental pathway for understanding cultural transformation and historical continuity, how MLLMs can be systematically leveraged to support and advance text evolution analysis remains an open and largely underexplored problem. To bridge this gap, we construct a comprehensive benchmark comprising 11 tasks and over 130,000 instances, specifically designed to evaluate the capability of MLLMs in analyzing the evolution of ancient Chinese scripts. We conduct extensive evaluations across multiple widely used MLLMs and observe that, while existing models demonstrate a limited ability in glyph-level comparison, their performance on core tasks—such as character recognition and evolutionary reasoning—remains substantially constrained. Motivated by these findings, we propose a glyph-driven fine-tuning framework (GEVO) that explicitly encourages models to capture evolutionary consistency in glyph transformations and enhances their understanding of text evolution. Experimental results show that even models at the 2B scale achieve consistent and comprehensive performance improvements across all evaluated tasks. To facilitate future research, we publicly release both the benchmark and the trained models\footnote{\url{https://github.com/songruiecho/GEVO}}.

\end{abstract}

\section{Introduction}
As carriers of Chinese cultural heritage, a deep understanding of ancient scripts plays a crucial role in preserving and promoting China’s rich cultural legacy~\cite{liu2025mcs}. In recent years, with the rapid advancement of Multimodal Large Language Models (MLLMs), a growing body of research has begun to leverage their capabilities to support the study of ancient scripts, ranging from character identification to critical interpretation, thereby demonstrating the significant potential of MLLMs in this domain~\cite{yao2025wenyangpt,qiao2025v,cao2025tonggu,li2025oracleagent}.

The understanding of ancient scripts is inherently tied to the analysis of character evolution, as writing styles from different historical periods exhibit intrinsic structural and semantic connections~\cite{bokset2006long}. Existing studies have explored the evolutionary analysis of ancient scripts, with particular attention to the transformation of character forms from oracle bone inscriptions to regular script~\cite{wang2022study,EVOBC2024,jiao2025graph}. However, as MLLMs become increasingly influential in related fields, systematic research on how to evaluate their capabilities in evolutionary analysis and how to effectively enhance these abilities remains limited.

To overcome the aforementioned gap, this paper comprehensively considers multiple datasets for the analysis of glyph evolution mentioned above, and reconstructs a dataset for the evolutionary analysis at the facsimile level, comprising 7,740 Chinese characters and nearly 30,000 corresponding facsimiles. Furthermore, to assess the capability of MLLMs in tasks related to the evolutionary analysis of ancient scripts, we construct a detailed evaluation benchmark based on the aforementioned dataset, with the assistance of paleography experts. This benchmark comprises three main tasks, 11 subtasks, and over 130,000 specific questions. 

Moreover, we evaluate the benchmark on 19 MLLMs ranging from 1B to 72B in scale and find that existing models exhibit limited capabilities in script style recognition cross-era and ancient script recognition, leading to weak fundamental performance in evolutionary analysis. Nonetheless, we find that with further fine-tuning on a small amount of data, MLLMs can exhibit substantial improvements in temporal attribution capability. Based on this observation, we gain confidence in the ability of MLLMs to handle evolutionary analysis tasks and propose a glyph-driven contrastive fine-tuning method inspired by curriculum learning to encourage models to fully discriminate the subtle differences induced by glyph forms and historical periods. Fine-tuning results on 2B-scale models confirm that the proposed method achieves consistent performance improvements across all tasks. In summary, our core contributions are as follows:

(i) We contribute a benchmark for evaluating the evolutionary analysis capabilities of MLLMs in Chinese ancient script. (ii) We evaluate this benchmark on multiple MLLMs, confirming that the evolutionary analysis capabilities of existing MLLMs are weak, but the recognition ability for script styles can be easily improved through fine-tuning. (iii) We propose a glyph-driven fine-tuning framework, which significantly enhances the performance of MLLMs. 

\section{Related Work}
\textbf{AI-Driven Research in Ancient Chinese Characters.} With the continuous advancement of algorithmic capabilities and computing resources, artificial intelligence has gradually permeated in the study of Chinese ancient scripts, which has significantly enhanced the efficiency of character interpretation, form comparison, and evolutionary analysis~\cite{wen2011chinese,li2026comprehensive}. This not only reduces repetitive labor for experts but also expands the boundaries of traditional research methods, offering new technological paradigms and research pathways for the development of related fields. Numerous benchmarks related to ancient Chinese characters, especially focusing on oracle bone inscriptions~\cite{GuanY0HLJBL24,wang2024open,zhou2025ancientbench}, have emerged. Given the pronounced long-tailed distribution and high noise levels characteristic of ancient scripts, research directions such as imbalanced learning~\cite{li2023decoupled,li2023towards,li2025mitigating}, denoising~\cite{shi2022charformer}, few-shot learning~\cite{zhao2022ffd}, and cross-modal learning~\cite{wang2024oracle} are also explored. 

\textbf{MLLM-Driven Research in Ancient Chinese Characters.} With the recent rapid development of MLLMs, related research has expanded from traditional deep learning to multi-task integration and generalization based on MLLMs. Some studies have proposed benchmarks for evaluating MLLMs, encompassing a variety of tasks related to ancient scripts, particularly oracle bone inscriptions. These benchmarks have demonstrated that even the most advanced models still lack the capability to handle such tasks effectively~\cite{chenobi,liu2025mcs}. Therefore, some approaches enhance the capability for text-related tasks by fine-tuning MLLMs on domain-specific datasets~\cite{cao2025tonggu}. There are also methods that leverage the core capabilities of MLLMs to assist experts in interpreting ancient scripts~\cite{qiao2025v}. Furthermore, some integrate models into agent frameworks, utilizing MLLMs to assist experts in tackling more complex tasks~\cite{li2025oracleagent}. Unlike previous studies, our primary contribution lies in constructing an evolutionary-related benchmark to evaluate the capabilities of MLLMs in evolution-related tasks. Additionally, we propose a glyph-based fine-tuning approach that effectively enhances model performance.

\section{Benchmark Construction}

\subsection{Construction of Glyph Evolution Dataset}
\begin{figure}[h]
    \centering
    \includegraphics[width=1\linewidth]{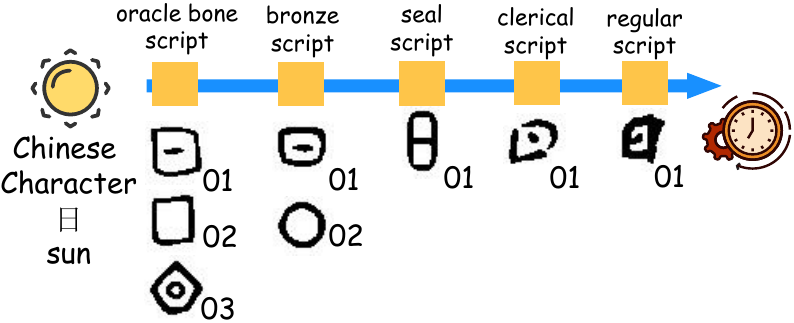}
    \caption{The text evolution process of the character ``\begin{CJK*}{UTF8}{gbsn}日\end{CJK*}'' (sun) and the organization of data.}
    \label{fig:example}
\end{figure}
\textbf{Construction of Basic Glyph Data.} Our dataset is considered based on existing evolutionary analysis datasets. We refer to the study~\cite{wang2022study} to divide the evolutionary process into five stages: Oracle Bone Script, Bronze Inscription, Seal Script, Clerical Script, and Regular Script. Subsequently, we apply the same stages to extract data from Vividict\footnote{https://www.vividict.com/} and extract the tracing data of characters in different periods into a unified organizational structure. As shown in Figure~\ref{fig:example}, this approach results in the same character potentially having multiple written forms within a single period, thereby enhancing the richness of the corpus. Subsequently, we uniformly binarize all font files and manually filter out instances of missing, corrupted, or blank data resulting from character extraction. This process ultimately yields a dataset encompassing the evolutionary processes of 7,740 characters. It is important to emphasize that not all characters possess a complete evolutionary path, which corresponds to the practical issue of missing evolutionary trajectories in paleographic research.
\begin{figure*}[h]
    \centering
    \includegraphics[width=1\linewidth]{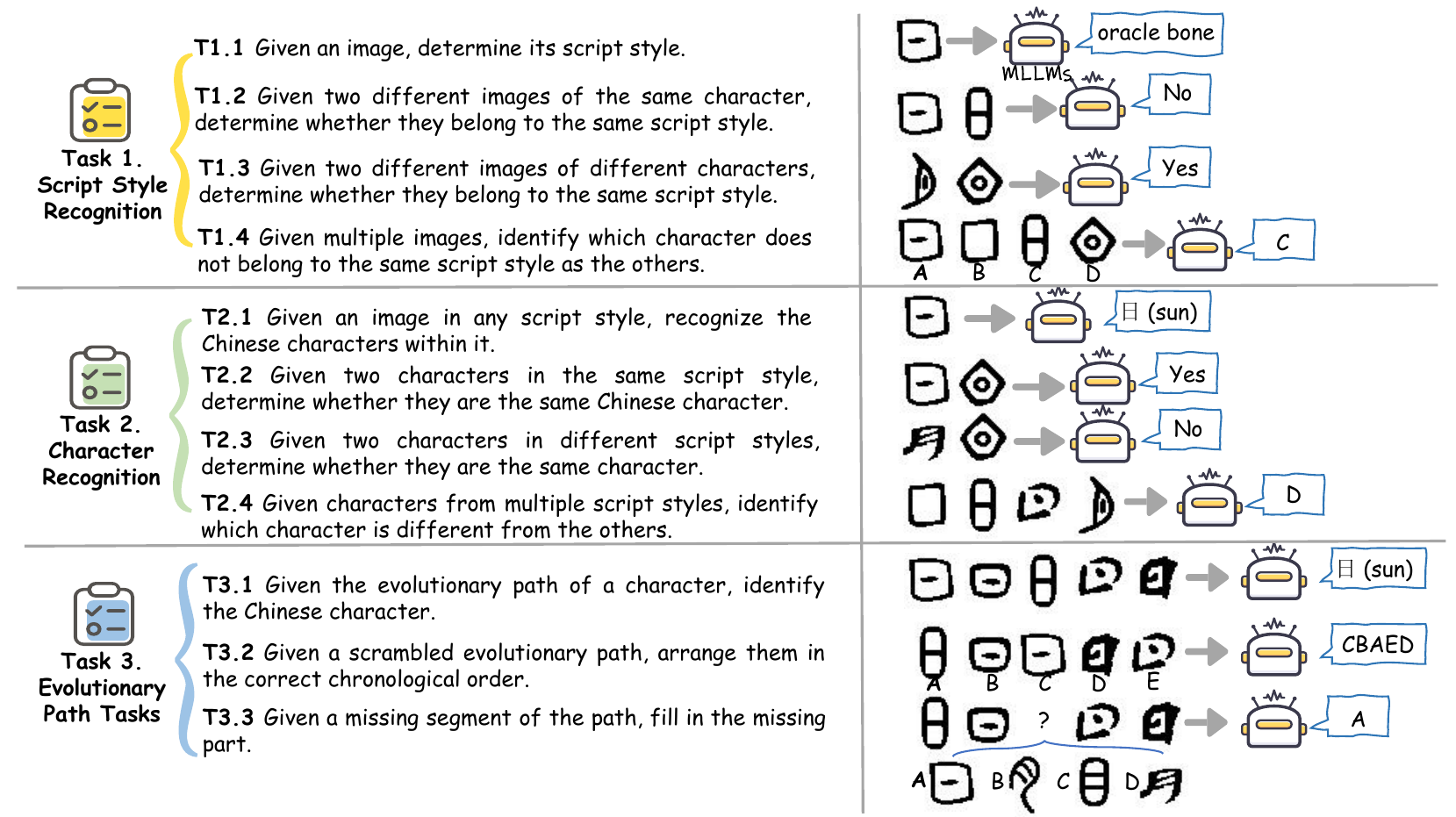}
    \caption{The 11 basic tasks that constitute the evaluation of MLLMs, along with corresponding examples. Due to limited space, we have provided a detailed expansion of the text instruction formats for different tasks in Appendix~\ref{app:task}.}
    \label{fig:tasks}
\end{figure*}
\textbf{Construction of Benchmark. } Subsequently, we aim to evaluate the evolution analysis capabilities of MLLMs. Therefore, with the assistance of paleographers, we abstract 11 subtasks into three major categories as shown in Figure~\ref{fig:tasks}. To adapt to MLLMs, the inputs for all tasks are modeled as a hybrid form of text instructions and images, with the expected output being text. Subsequently, we design task-specific instruction prompts tailored to different MLLMs, framing the tasks in formats such as question answering, binary judgment, and multiple-choice, thereby constructing the final benchmark. To enable the benchmark to better align with the fine-tuning–evaluation paradigm, we randomly split each task into training and test sets by 9:1. 

\begin{figure}[t]
    \centering
    \includegraphics[width=1\linewidth]{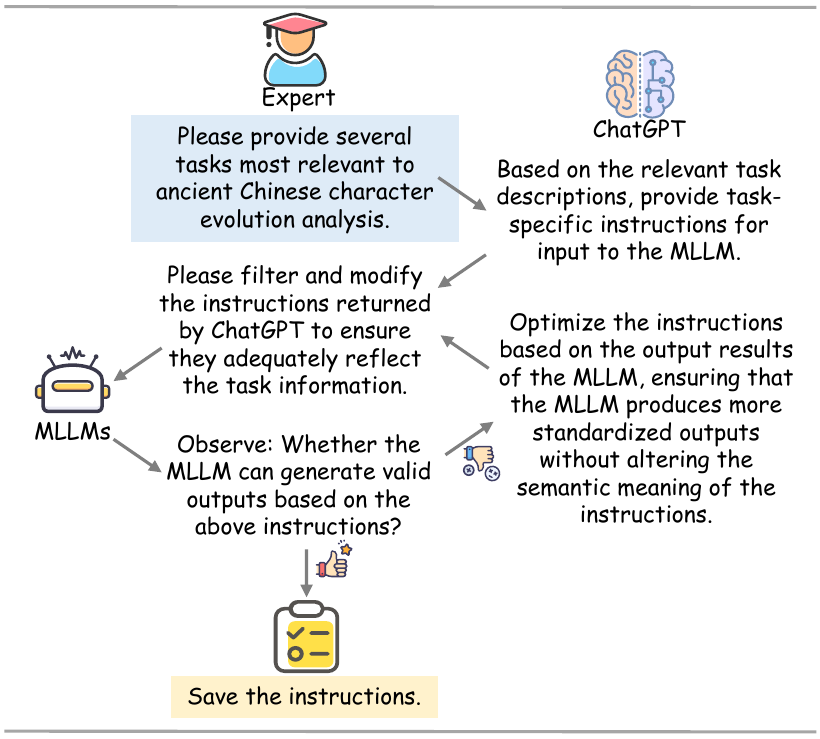}
    \caption{Task-related instruction construction and optimization process.}
    \label{fig:process}
\end{figure}
Specifically, during the instruction construction process, we consulted paleographers to define the general task scope. Subsequently, based on this task scope, we solicit suggestions from ChatGPT to generate a set of candidate instructions, which were then presented again to paleography experts to ensure professionalism. These instructions were subsequently input into multiple locally deployable MLLMs (the \texttt{Qwen3} series and \texttt{InternVL} series evaluated in this paper) to verify their ability to guide the MLLMs in producing standardized answers and reasonable outputs. For instructions that proved ineffective, we revise them based on expert feedback and MLLM responses, ultimately ensuring the task validity and model compliance of the instructions. This process is illustrated in Figure~\ref{fig:process}.

\subsection{Dataset Statistics}
\begin{figure}[t]
    \centering
    \begin{subfigure}[t]{0.49\linewidth}
        \centering
        \includegraphics[width=\linewidth]{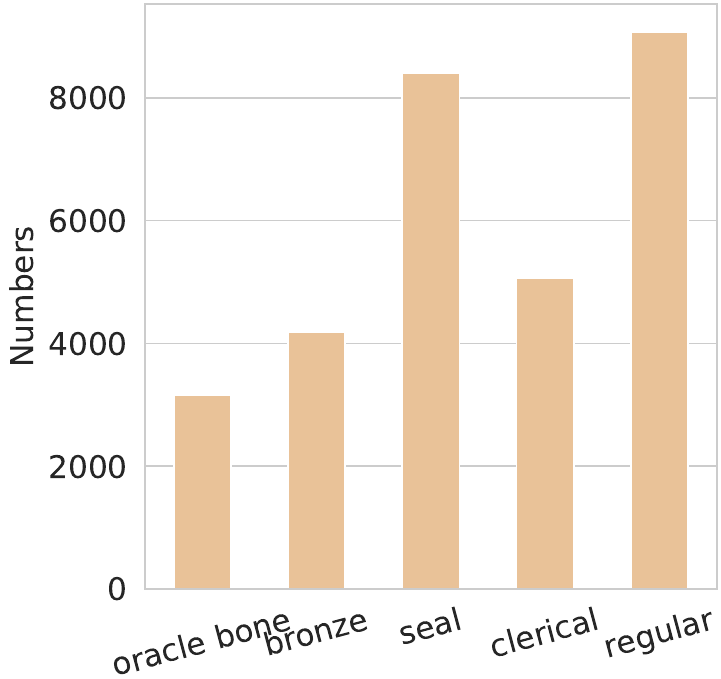}
        \caption{Quantity of different script styles.}
        \label{fig:n1}
    \end{subfigure}
    \hfill
    \begin{subfigure}[t]{0.49\linewidth}
        \centering
        \includegraphics[width=\linewidth]{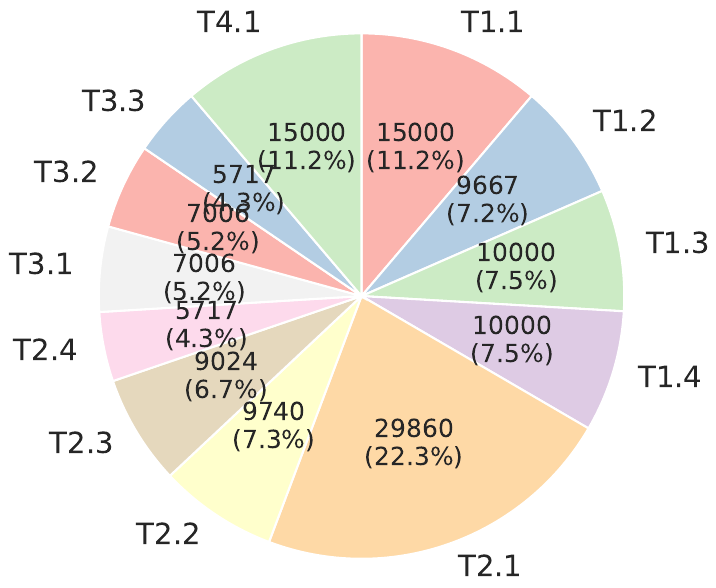}
        \caption{Distribution of benchmark questions.}
        \label{fig:n2}
    \end{subfigure}
    \caption{Benchmark quantity statistics.}
    \label{fig:n}
\end{figure}
Figure~\ref{fig:n} illustrates the data distribution of the benchmark across different ancient scripts and their corresponding tasks. Oracle bone inscriptions account for the smallest portion of the dataset, as only slightly more than two thousand oracle bone characters have been deciphered and can be reliably mapped to modern Chinese characters~\cite{huang2019obc306}. In contrast, regular script contains the largest number of images, since it is the closest in form to modern Chinese characters and thus more abundantly preserved. From a task perspective, \textbf{T2.1} includes the largest number of instances, as it is designed to comprehensively evaluate the image-level recognition capabilities of MLLMs. Conversely, \textbf{T3.3} has the smallest data volume because it requires a complete and well-documented evolutionary trajectory for each character, which is unavailable for many characters. Overall, the benchmark is constructed at a moderate and well-balanced scale, enabling a comprehensive and multi-dimensional assessment of MLLMs.

\section{Evaluation of MLLMs}
\begin{table*}[t]
  \centering
  \small
  \setlength\tabcolsep{3.5pt}
  \renewcommand{\arraystretch}{0.9}
    \begin{tabular}{c|cccc|cccc|ccc|c}  \toprule
    MLLMs& T1.1  & T1.2  & T1.3  & T1.4  & T2.1  & T2.2  & T2.3  & T2.4  & T3.1  & T3.2  & T3.3  & Average \\ \midrule
    \rowcolor[rgb]{ .859,  .859,  .859} \multicolumn{13}{c}{\texttt{Closed-source Models}} \\
    \texttt{GPT4-all-mini} & 30.85  & 51.09  & 48.30  & 4.00  & 0.30  & 50.10  & 49.17  & 1.05  & 0.00  & 19.23  & 24.13  & 25.29  \\
    \texttt{GPT5-mini} & 24.65  & 46.84  & 46.30  & 11.50  & 0.07  & 49.50  & 48.70  & 6.79  & 0.00  & 13.60  & 25.78  & 24.88  \\
    \texttt{Gemini-3-Flash} & 21.73  & 48.71  & 48.90  & 24.20  & 0.03  & 48.46  & 52.27  & 25.17  & 3.74  & 12.06  & 21.52  & 27.89  \\
    \rowcolor[rgb]{ .859,  .859,  .859} \multicolumn{13}{c}{\texttt{Open-source Models}} \\
    \texttt{TongGu-VL-2B-Instruct} & 4.07  & 50.78  & 52.90  & 15.00  & 8.64  & 49.79  & 52.49  & 1.54  & 9.27  & 4.39  & 26.57  & 25.04  \\
    \texttt{Qwen2.5-VL-7B-Instruct} & 31.07  & 63.91  & 57.60  & 38.70  & 23.51  & 75.46  & 74.20  & 42.83  & 40.66  & 36.40  & 39.86  & 47.65  \\
    \texttt{Qwen2.5-VL-32B-Instruct} & 24.00  & 55.33  & 56.80  & 29.90  & 23.58  & 71.25  & 74.64  & 50.00  & 39.80  & 27.72  & 56.29  & 46.30  \\
    \texttt{Qwen2.5-VL-72B-Instruct} & 18.60  & 61.49  & 57.30  & 41.80  & 24.45  & 73.31  & 74.86  & 54.72  & 49.79  & 48.88  & 62.76  & 51.63  \\
    \texttt{Qwen3-VL-2B-Instruct} & 17.81  & 55.74  & 51.90  & 26.70  & 21.23  & 70.74  & 71.43  & 22.73  & 27.39  & 16.67  & 44.93  & 38.84  \\
    \texttt{Qwen3-VL-8B-Instruct} & 45.27 & 76.63  & 69.70  & 56.80  & 30.74  & 74.13  & 81.62  & 69.93  & 49.36  & 50.43  & 71.68  & 61.48  \\
    \texttt{Qwen3-VL-30B-A3B-Instruct} & 32.81  & 66.94  & 58.02  & 45.37  & 31.12  & 70.89  & 77.45  & 63.84  & 55.91  & 36.02  & 64.28  & 54.79  \\
    \texttt{InternVL3\_5-1B-HF} & 7.60  & 53.15  & 51.20  & 21.90  & 17.35  & 46.71  & 43.08  & 11.71  & 22.11  & 7.57  & 33.04  & 28.67  \\
    \texttt{InternVL3\_5-8B-HF} & 17.33  & 58.53  & 52.00  & 28.01  & 4.69  & 57.49  & 54.37  & 36.19  & 15.98  & 18.98  & 40.21  & 37.99  \\
    \texttt{InternVL3\_5-14B-HF} & 7.33  & 48.50  & 48.30  & 33.90  & 2.55  & 55.95  & 54.04  & 74.48  & 18.69  & 33.51  & 74.30  & 41.05  \\
    \texttt{MiniCPM-V-2\_6 8B} & 7.67  & 55.02  & 51.90  & 28.10  & 8.84  & 80.70  & 76.41  & 45.98  & 7.85  & 16.00  & 25.87  & 36.76  \\
    \texttt{MiniCPM-V-4\_5 8B} & 21.00  & 62.67  & 54.10  & 36.00  & 9.31  & 80.49  & 77.41  & 64.34  & 13.84  & 25.83  & 25.70  & 42.79  \\
    \texttt{GLM-4.1V-9B-Thinking} & 26.00  & 52.43  & 51.00  & 23.20  & 18.05  & 46.30  & 43.30  & 11.36  & 20.68  & 7.73  & 34.09  & 30.38  \\
    \texttt{DeepSeekOCR-3B} & 11.12  & 50.34  & 48.21  & 25.87  & 9.64  & 51.02  & 56.03  & 21.18  & 18.35  & 15.72  & 45.08  & 32.05  \\
    \texttt{LLaVA-1.5-7B-HF} & 10.47  & 9.20  & 8.40  & 26.10  & 0.00  & 22.98  & 24.92  & 22.90  & 0.00  & 12.36  & 26.22  & 14.87  \\
    \texttt{LLaVA-1.5-13B-HF} & 3.33  & 47.57  & 50.10  & 11.50  & 0.17  & 59.24  & 61.46  & 5.94  & 0.00  & 6.50  & 28.96  & 24.98  \\
     \texttt{GEVO}*  & 80.60  & 88.83  & 87.00  & 83.20  & 39.18  & 89.32  & 92.03  & 96.85  & 70.19  & 93.13  & 98.60  & 83.54  \\ 
    \bottomrule
    \end{tabular}%
    \caption{Performance comparison of different MLLMs across all test sets (accuracy \%). The evaluation results on the full dataset are provided in Appendix~\ref{app:eval}. } 
  \label{tab:eval}
\end{table*}

To validate the capability of existing MLLMs in tasks related to evolutionary analysis, we employ 19 commonly used MLLMs to evaluate all of the aforementioned tasks on test sets as shown in Table~\ref{tab:eval}. Based on the experimental results, we provide the most important observations as follows:

\textbf{MLLMs have a certain ability to compare script styles and character glyphs, but its character recognition capability is relatively poor.} Compared to script style recognition (T1.1), the performance of text recognition tasks (T2.1) is generally lower. This is because character recognition requires strong expert knowledge, and the existing models often lack this specialized expertise. When tasks are concretized to script style (T1.1, T1.2, T1.3) and glyph comparison and selection (T2.1, T2.2, T2.3), the performance often improves compared to simple recognition tasks. This indicates that, rather than expert knowledge-driven specialized recognition tasks, MLLMs are more adept at distinguishing similarities and differences through comparison. Additionally, in some cases, such as the \texttt{LLaVA-1.5} series models, their performance is poor due to difficulties in understanding complex instructions in ancient script research. Additionally, for some closed-source models, they exhibit almost no character recognition capability or frequently refuse to provide valid responses, leading to an average performance that is often inferior to that of local models.

\textbf{Compared with isolated glyph comparison and recognition, performing the same tasks within an explicit evolutionary context leads to improved performance, underscoring the importance of modeling the evolutionary process.} In most cases, character recognition (T3.1) and script-style identification (T3.2) benefit from being situated within the evolutionary process, as the model can leverage additional contextual information to support reasoning and evaluation. This observation highlights the necessity of incorporating evolutionary knowledge into ancient script analysis. Nevertheless, owing to the limited domain-specific knowledge of MLLMs in ancient script studies, the observed performance gains remain relatively modest. For T3.3, performance largely depends on the model’s ability to jointly compare glyph forms and semantic meanings; as a result, accuracy in many cases approaches random guessing (25\%). This further emphasizes the need to equip MLLMs with more specialized capabilities for ancient script understanding.

\subsection{Analysis of Results}
\begin{figure}[t]
    \centering
    \includegraphics[width=1\linewidth]{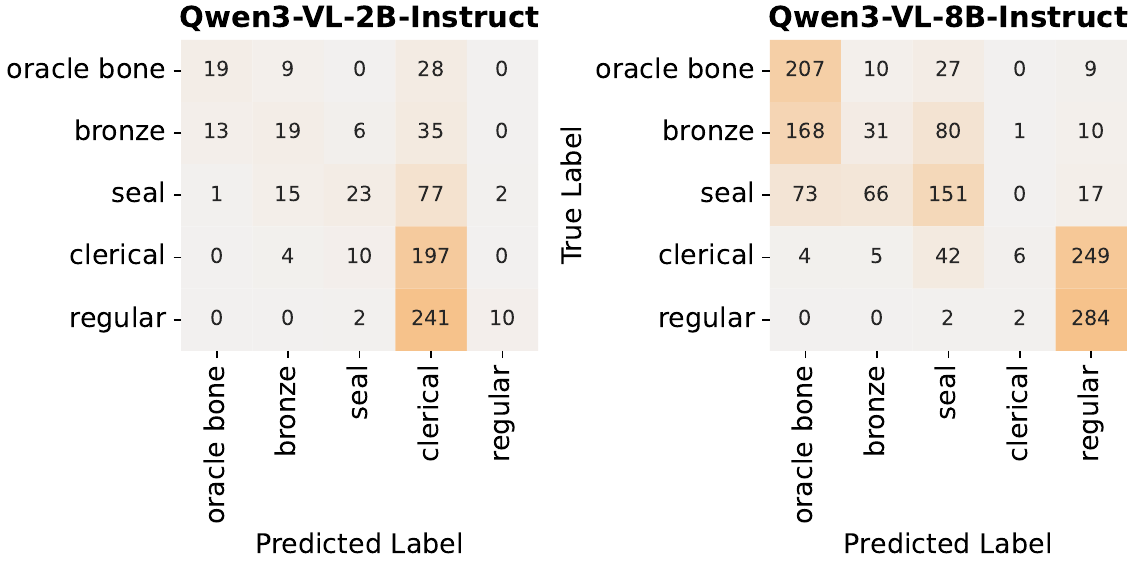}
    \caption{The confusion matrices of script style predictions for different models, with unrecognizable cases removed to better highlight the relationships among glyphs across scripts and periods.}
    \label{fig:conf}
\end{figure}

\begin{figure}[t]
    \centering
    \includegraphics[width=1\linewidth]{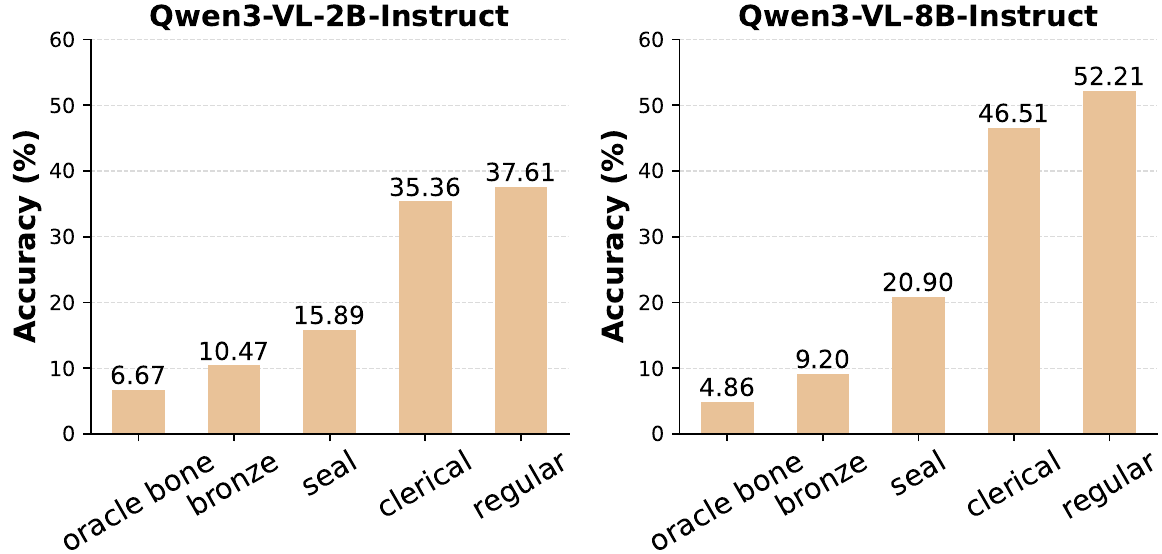}
    \caption{Characters accuracy on different script styles.}
    \label{fig:style_acc}
\end{figure}
Based on the quantitative results above, we aim to further investigate how script style influences predictions. Given that the \texttt{Qwen} series of models demonstrate the most outstanding performance on the evaluation benchmark, we further conduct an in-depth analysis of the prediction results from the \texttt{Qwen3-VL-2B-Instruct} and \texttt{Qwen3-VL-8B-Instruct} in Figure~\ref{fig:conf} and \ref{fig:style_acc}. 

Figure~\ref{fig:conf} illustrates the confusion in script styles across adjacent historical periods, particularly among oracle bone, bronze, and seal script, as well as between clerical script and regular script. This observation is well aligned with the principles of script evolution, as scripts from adjacent periods tend to exhibit highly similar stylistic characteristics. As for the sparsity observed in \texttt{Qwen3-VL-2B-Instruct}'s performance on oracle bone, bronze, and seal scripts, it occurs because the model tends to "decline to answer" when it is uncertain. This directly demonstrates the limitations in the capability of a 2B-scale model.

Figure~\ref{fig:style_acc} reports the prediction accuracy across different script styles. A clear and intuitive trend can be observed: scripts that are closer to the modern era consistently achieve higher recognition accuracy. This pattern is consistent with the evolutionary trajectory of Chinese characters, which gradually converge toward modern forms, thereby enabling MLLMs to more effectively transfer their intrinsic knowledge of contemporary Chinese characters. In contrast, the recognition accuracy for oracle bone script remains below 10\%, indicating that current MLLMs possess little to no effective capability for recognizing this script. This observation further underscores the necessity of endowing MLLMs with specialized knowledge to support research on ancient scripts.

\subsection{Preliminary Attempt: Few-shot SFT}
\begin{figure}[t]
    \centering
    \includegraphics[width=1\linewidth]{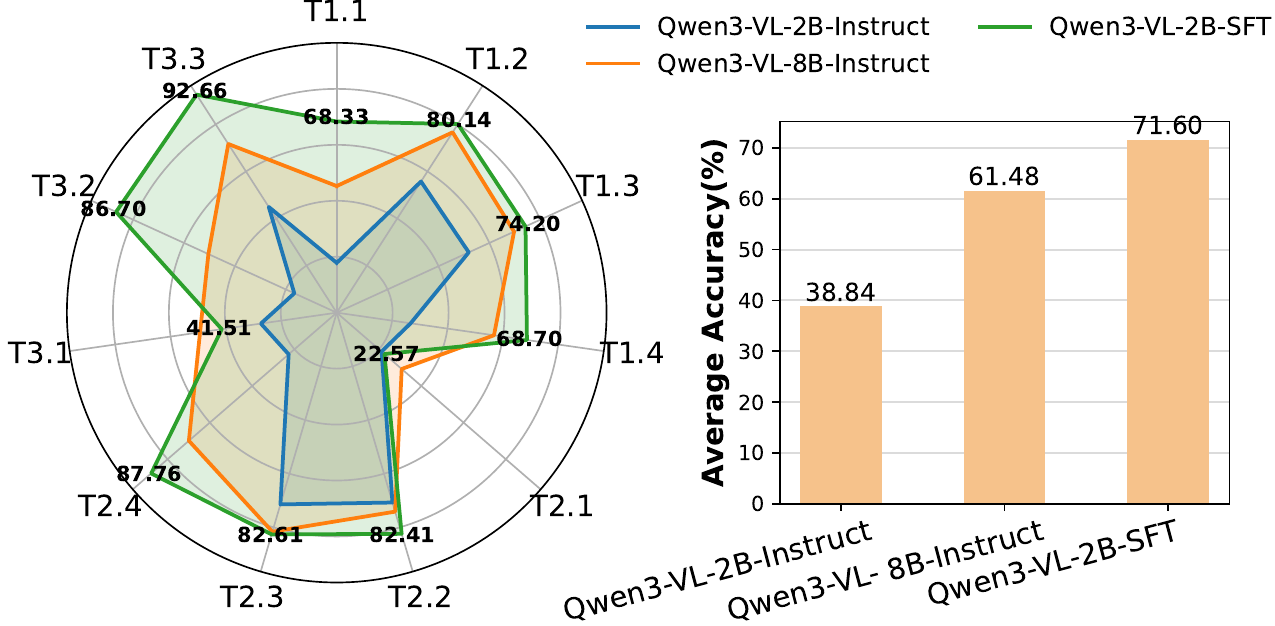}
    \caption{Comparison of results for \texttt{Qwen3-VL-2B-Instruct} after SFT (\texttt{Qwen3-VL-2B-SFT}) across all tasks.}
    \label{fig:sft}
\end{figure}
To investigate the learning potential of MLLMs for tasks related to ancient text evolution analysis, we adopt \texttt{Qwen3-VL-2B-Instruct} and conduct simple supervised fine-tuning (SFT) using 200 randomly sampled training examples per task. The SFT results are summarized in Figure~\ref{fig:sft}. Overall, the fine-tuned model, \texttt{Qwen3-VL-2B-SFT}, exhibits substantial performance gains, with an average improvement exceeding 30\% over the original model, and even surpasses 8B-scale models by more than 10\%. These results suggest that MLLMs can be effectively adapted to glyph comparison tasks for ancient scripts, as tracings of ancient characters are generally not visually complex.

However, we also observe performance degradation of \texttt{Qwen3-VL-2B-SFT} on T2.1 and T3.1. This indicates that a limited number of training samples is insufficient to support robust recognition of ancient scripts, and may even induce catastrophic forgetting of previously acquired knowledge. Taken together, these findings reveal two key principles for guiding MLLM training: \textbf{(i) a small number of samples can suffice to enhance the model’s ability to discriminate similar characters; and (ii) comprehensive and sufficiently diverse data are essential to endow the model with stable and reliable recognition capabilities for writing systems from temporally distant historical periods.}

\section{Glyph-driven Curriculum Learning}

\subsection{Model Framework}
\newcommand{\glyph}[2][2.5ex]{%
  \raisebox{-0.3ex}{\includegraphics[height=#1]{#2}}%
}
\begin{figure}[t]
    \centering
    \includegraphics[width=1\linewidth]{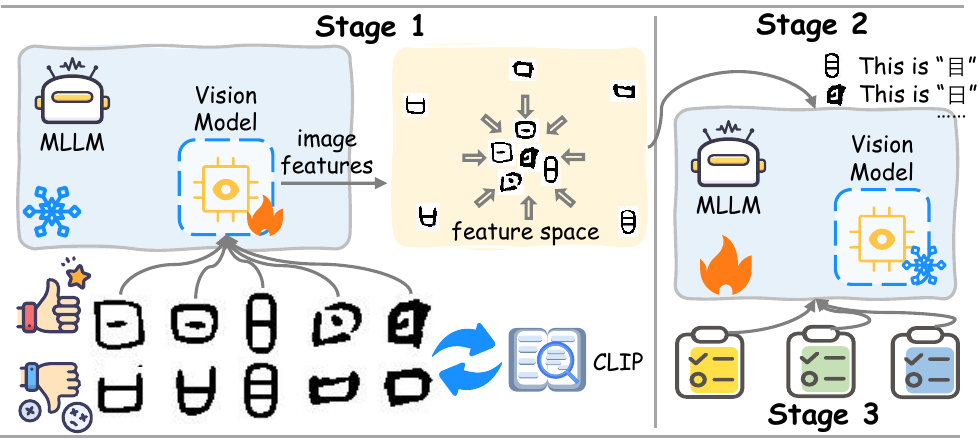}
    \caption{The two-stage framework of GDEVA training. In the first stage, characters in different script styles that represent the same character ``\begin{CJK*}{UTF8}{gbsn}日\end{CJK*}" (sun) are treated as a set of positive samples, while other images that are glyphically similar to them are considered negative samples. For example, \glyph{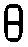} and \glyph{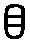} are considered glyphically similar, but they are different ways of writing the characters for ``\begin{CJK*}{UTF8}{gbsn}日\end{CJK*}" (sun) and ``\begin{CJK*}{UTF8}{gbsn}目\end{CJK*}" (eye), respectively. Therefore, during the training process, we need to maintain a distance between them. }
    \label{fig:model}
\end{figure}
Inspired by the results of our preliminary experiments, we propose a multi-stage fine-tuning framework as shown in Figure~\ref{fig:model}. The framework is built upon curriculum learning, where the model progressively learns to understand evolutionary processes by tackling tasks of increasing complexity, from simple to difficult~\cite{wang2021survey}. \underline{In the first stage}, we aim to model the glyph variations of the same character across different script styles, while enhancing the alignability for visual representation. To this end, we independently fine-tune the visual module of MLLM, specifically updating the parameters of the visual encoder and the cross-modal projection module, so as to learn multimodal representations that are discriminative with respect to glyph variations while remaining semantically consistent.

Specifically, we employ a contrastive learning approach~\cite{chen2020simple} to encourage the representation optimization of the vision model within the MLLM. For each character, its script style-related images $I_n$ corresponding to different historical periods are regarded as a set of positive samples $\mathcal{P}=\{I_1,...I_n\}$, as they all represent the same character. Additionally, to mitigate interference from visually similar glyphs of different characters, we employ CLIP~\cite{radford2021learning} to retrieve the top-$k$ most visually similar glyph images that do not correspond to the target character, which are then used to construct negative samples $\mathcal{N}=\{\neg I_1^{(1)},..., \neg I_1^{(k)}, ..., \neg I_n^{(1)}, ..., \neg I_n^{(k)}\}$. Here, $\neg I_1^{(1)}$ represents the first negative sample of $I_1$, with a maximum of $k$. Subsequently, the following contrastive learning loss is optimized to encourage the model to learn the similarities and differences between different images:

\begin{equation}
\mathcal{L}_{con}
= - \frac{1}{|\mathcal{P}|}
\sum_{I_i \in \mathcal{P}}
\log
\frac{\mathcal{S}_i^+}{\mathcal{S}_i^+ + \mathcal{S}_i^-},
\end{equation}
where $\mathcal{S}_i^+ = \sum\limits_{{I_j \in \mathcal{P}, I_j \neq I_i}} e^{\frac{s(\mathbf{z}_i, \mathbf{z}_j)}{\tau}}$ and $\mathcal{S}_i^- = \sum\limits_{I^- \in \mathcal{N}_i}
e^{\frac{s(\mathbf{z}_i, \mathbf{z}^-)}{\tau}}$. Here, $s(\mathbf{z}_i, \mathbf{z}_j)$ is cosine similarity between image representations, and $\mathbf{z}_i, \mathbf{z}^-$ represent the representations of $I_i$ and its corresponding negative sample, respectively. 

\underline{In the second stage}, we aim for the model to further learn the mapping between images and text based on the glyphs it has already learned. Specifically, given an image of a character from any historical script period, the model should predict the corresponding modern Chinese character. During this process, the parameters of the language model in the MLLMs are updated and keep the visual model parameters frozen, primarily capturing semantic associations.

\underline{In the third stage}, we fine-tune the language model in the MLLMs using task-related instructions. Similarly, SFT is performed on a dataset containing only 200 samples per task to reduce the cost of fine-tuning. We name this glyph-driven evolutionary MLLM as \textbf{GEVO}. More details regarding model fine-tuning can be found in Appendix~\ref{app:sft}. 

\subsection{Result Evaluation} 
\begin{table*}[h]
  \small
  \centering
  \setlength\tabcolsep{3.5pt}
  \renewcommand{\arraystretch}{0.9}
    \begin{tabular}{c|cccc|cccc|ccc|c}  \toprule
    MLLMs& T1.1  & T1.2  & T1.3  & T1.4  & T2.1  & T2.2  & T2.3  & T2.4  & T3.1  & T3.2  & T3.3  & Average \\ \midrule
    \texttt{Qwen3-VL-2B-SFT} & 68.33  & 80.14  & 74.20  & 68.70  & 22.57  & 82.41  & 82.61  & 87.76  & 41.51  & 86.70  & 92.66  & 71.60  \\
    \texttt{GEVO-Stage1} & 75.00  & 83.14  & 78.70  & 78.10  & 1.64  & 62.01  & 47.84  & 26.75  & 4.42  & 90.13  & 93.65  & 58.31  \\
    \texttt{GEVO-Stage2} & 32.20  & 57.19  & 50.40  & 32.40  & 28.20  & 81.62  & 89.37  & 92.13  & 54.64  & 33.12  & 95.45  & 58.79  \\
    \texttt{GEVO}*  & 80.60  & 88.83  & 87.00  & 83.20  & 39.18  & 89.32  & 92.03  & 96.85  & 70.19  & 93.13  & 98.60  & 83.54  \\ \bottomrule
    \end{tabular}%
  \caption{Performance comparison of different GEVO variants across all tasks. * indicates a significant performance improvement under Wilcoxon Signed-Rank Test ($p<0.05$).}
  \label{tab:result_sft}%
\end{table*}

Table~\ref{tab:result_sft} reports a comparative evaluation of different GEVO variants. Specifically, \texttt{GEVO-Stage1} refers to the setting in which the model, after completing training in Stage~1, proceeds directly to supervised fine-tuning in Stage~3, while \texttt{GEVO-Stage2} denotes the variant in which glyph-based contrastive learning in Stage~1 is omitted and fine-tuning is conducted only in Stages~2 and~3. 

The experimental results demonstrate that, relative to directly fine-tuning \texttt{Qwen3-VL-2B-SFT}, both GEVO variants experience a clear degradation in overall performance. The results for \texttt{GEVO-Stage1} indicate that glyph-driven contrastive learning effectively enhances performance on glyph-centric tasks, particularly the T1 series and T3.2. This observation further underscores the critical role of glyph information in determining script style and historical period, and suggests that task designs emphasizing glyph comparison can yield tangible performance gains. However, \texttt{GEVO-Stage1} exhibits a pronounced decline on character recognition tasks (the T2 series) as well as T3.1, with accuracy on T2.1 and T3.1 dropping below 10\%. This behavior indicates that training focused exclusively on glyph-level signals induces catastrophic forgetting of the model’s already limited character recognition capabilities. In contrast, the results for \texttt{GEVO-Stage2} show that emphasizing recognition-oriented training improves character identification, leading to notable gains on T2.1 and T3.1. Nevertheless, these gains come at the cost of a severe degradation in glyph comparison performance, leading to a collapse in effectiveness on glyph-related tasks.

GEVO effectively balances the model’s glyph comparison and character recognition capabilities, yielding substantial improvements across all tasks. Even when compared with \texttt{Qwen3-VL-2B-SFT}, GEVO achieves an average performance gain exceeding 10\%. Moreover, it outperforms the baseline by more than 10\% on both fundamental tasks, T1.1 and T2.1, demonstrating the effectiveness of the training methodology underlying GEVO. By simultaneously enhancing glyph-level discrimination while preserving character recognition ability, this approach establishes stronger foundational competencies, which in turn translate into improved performance across a broader range of downstream tasks. But it is important to note that GEVO’s character recognition performance (39.18\%) remains relatively limited, indicating that substantial room for further improvement still exists in enhancing the character recognition capabilities of MLLMs.

\subsection{Further Analysis}

\begin{figure}[t]
    \centering
    \includegraphics[width=1\linewidth]{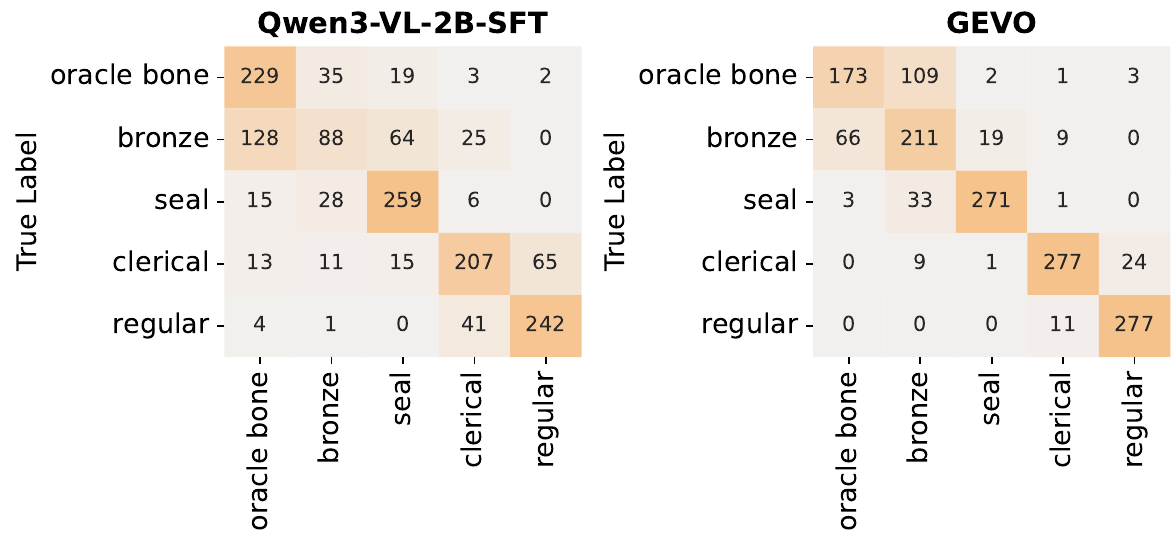}
    \caption{The confusion matrices of script style predictions for \texttt{Qwen3-VL-2B-SFT} and GEVO.}
    \label{fig:confusion_gevo}
\end{figure}

\begin{figure}[t]
    \centering
    \includegraphics[width=1\linewidth]{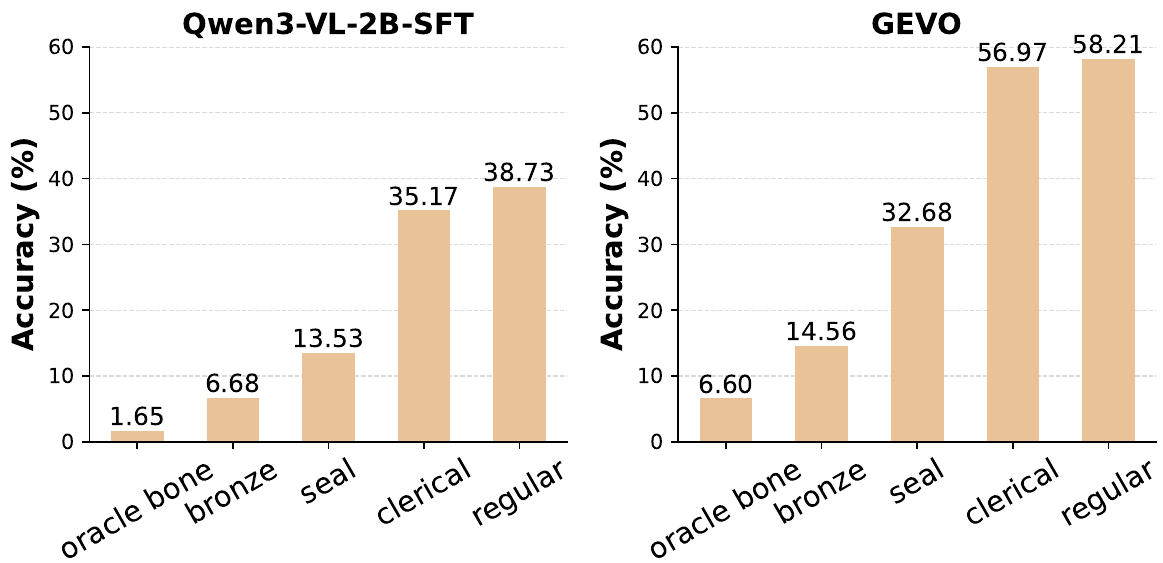}
    \caption{Accuracy of characters across different script styles for \texttt{Qwen3-VL-2B-SFT} and GEVO.}
    \label{fig:style_acc_gevo}
\end{figure}

Figure~\ref{fig:confusion_gevo} presents the confusion matrix for script style recognition. Although oracle bone script and bronze script still exhibit a certain degree of confusion due to their high glyph-level similarity, the overall prediction quality is substantially improved. Compared with Figure~\ref{fig:conf}, the number of correctly and reasonably predicted instances increases markedly, and the model no longer fails to respond to specialized queries. In addition, the confusion between seal script and clerical script is significantly reduced. These results indicate that SFT on related tasks can effectively enhance the MLLM’s understanding and discrimination capability for ancient script–related tasks. Furthermore, when comparing the two SFT variants, \texttt{Qwen3-VL-2B-SFT} and GEVO, we find GEVO further mitigates the confusion between clerical script and regular script observed in \texttt{Qwen3-VL-2B-SFT}, while simultaneously enhancing the predictive performance on bronze script. This suggests that GEVO effectively benefits from explicit glyph-level comparisons. 

Figure~\ref{fig:style_acc_gevo} reports the character recognition accuracy across different script styles. Compared with the results of \texttt{Qwen3-VL-2B-Instruct} shown in Figure~\ref{fig:style_acc}, \texttt{Qwen3-VL-2B-SFT} exhibits performance gains only on regular script. This suggests that a limited amount of training data is insufficient to substantially improve character recognition, as the pronounced glyph variations across different historical periods of Chinese characters cannot be effectively generalized from a small number of samples. Consequently, ancient character recognition is inherently a knowledge-intensive task, which partly explains the weak performance of general-purpose models in this domain. Furthermore, \texttt{Qwen3-VL-2B-SFT} shows degraded recognition performance on oracle bone script and bronze script, indicating that MLLMs tend to exhibit representational bias toward script styles with more stable structures and regular strokes. Such bias weakens the models’ ability to capture the highly heterogeneous and non-standard glyph forms characteristic of early scripts. By incorporating character recognition training tasks, GEVO achieves substantial performance improvements, with accuracy exceeding 50\% on both clerical script and regular script. However, the gains on oracle bone script and bronze script remain limited. This observation indicates that, even with fine-tuning, accurate recognition of ancient scripts remains challenging for MLLMs, highlighting the need for more comprehensive and efficient datasets to facilitate deeper and more robust learning.

\subsection{Visualization Analysis} \label{sec:vis}
\begin{figure}[h]
    \centering
    \includegraphics[width=1\linewidth]{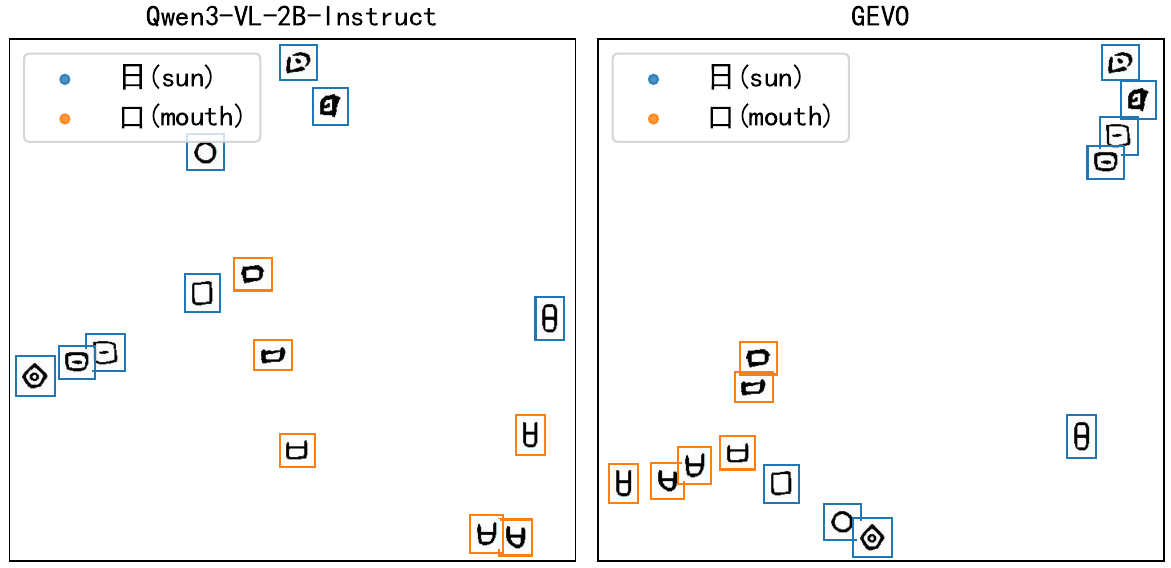}
    \caption{Visualization of similar image representations among ``\begin{CJK*}{UTF8}{gbsn}日\end{CJK*}'' (sun) and ``\begin{CJK*}{UTF8}{gbsn}口\end{CJK*}'' (mouth) baed on \texttt{Qwen3-VL-2B-Instruct} and GEVO in two-dimensional space. Boxes in different colors are used to distinguish images of script styles corresponding to different modern Chinese characters.}
    \label{fig:vis_ri}
\end{figure}

\begin{table*}[h]
\centering
\small
\setlength{\tabcolsep}{5pt}
\begin{tabular}{cccccccccc|c}
\toprule
\textbf{MLLMs} 
& \textbf{T1.1} & \textbf{T1.3} & \textbf{T1.4} 
& \textbf{T2.1} & \textbf{T2.3} & \textbf{T2.4} 
& \textbf{T3.1} & \textbf{T3.2} & \textbf{T3.3} 
& \textbf{Average} \\
\midrule
\texttt{GEVO}
& 69.55 & 81.95 & 79.80 
& 56.84 & 87.84 & 92.57 
& 90.54 & 79.75 & 97.30 
& 81.79 \\

\texttt{Qwen3-VL-2B-Instruct}
& 24.30 & 59.05 & 28.50 
& 41.34 & 69.59 & 25.00 
& 70.27 & 23.09 & 46.62 
& 43.08 \\

\texttt{Qwen3-VL-8B-Instruct}
& 45.53 & 70.55 & 61.20 
& 47.63 & 75.34 & 65.54 
& 71.62 & 45.62 & 74.32 
& 61.93 \\
\bottomrule
\end{tabular}
\caption{Performance comparison on different tasks.}
\label{tab:ood_results}
\end{table*}

We conduct a visualization analysis to explore GEVO’s ability to distinguish glyphs in the representation space, thereby providing substantial evidence for its performance in downstream tasks. For this purpose, Figure~\ref{fig:vis_ri} presents the representation distribution of the two distinct characters ``\begin{CJK*}{UTF8}{gbsn}日\end{CJK*}'' (sun) and ``\begin{CJK*}{UTF8}{gbsn}口\end{CJK*}'' (mouth). Since both are pictographic characters derived from real-world objects, they are often compressed into a limited number of stable geometric prototypes during the early stages, leading to convergent geometric abstraction at the level of outer contours. For more intuitive visualization, different images are positioned in the corresponding two-dimensional space based on their representations. Among them, GEVO demonstrates superior clustering ability for identical characters, positioning different styles of the character ``\begin{CJK*}{UTF8}{gbsn}口\end{CJK*}'' (mouth) at similar distances. Additionally, the representations of the glyphs for the same character ``\begin{CJK*}{UTF8}{gbsn}日\end{CJK*}'' (\glyph{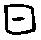} \glyph{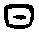} \glyph{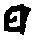} \glyph{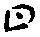}) have also become more concentrated. But we still observe that the model struggles to distinguish particularly similar glyphs, such as \glyph{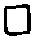} and \glyph{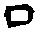}. This indicates that GEVO still has room for improvement, particularly in distinguishing especially similar glyphs. We provide more interesting visualization results with greater semantic differences in Appendix~\ref{app:vis}.

\section{Generalization on OOD Datasets}
Given the potential discrepancies among glyph facsimiles from different sources, we further extract a Out-of-distribution (OOD) subset from OBIsEvolution~\cite{wang2022study} for evaluating the generalization ability of GEVO. Specifically, we randomly select 150 characters from the dataset and manually filtered out questionable samples, resulting in a small-scale dataset containing 148 characters and 717 corresponding facsimiles. Subsequently, following the same procedure adopted in GEVO, we constructed an MLLM evaluation instruction set and re-evaluated GEVO’s inference results on the newly constructed benchmark. It should be emphasized that tasks \textbf{T1.2} and \textbf{T2.2} can not be conducted in this OOD benchmark, as each glyph has only a single version and thus lacks multiple variants for evaluation.

The experimental results in Table~\ref{tab:ood_results} show that GEVO consistently outperforms the baseline MLLMs across all evaluation tasks, demonstrating superior robustness and generalization ability in out-of-distribution settings. Compared with general-purpose MLLMs, GEVO achieves more stable and balanced performance on diverse subtasks, indicating its stronger capability in handling challenging ancient character reasoning scenarios. These results validate the effectiveness of the proposed framework in enhancing ancient character understanding under distribution shifts.

\section{Conclusion}
This paper introduces a benchmark for evaluating MLLMs on Chinese character evolution tasks. Evaluations of 19 MLLMs reveal persistent weaknesses in glyph comparison and character recognition, though modest gains can be achieved through simple SFT. Bassed on the findings, we propose a curriculum-inspired fine-tuning approach based on glyph contrastive learning, which improves performance across tasks. Notably, fine-tuned 2B-scale models surpass all evaluated MLLMs. 

\section*{Limitations}
Since ink rubbings often contain noise and exhibit highly inconsistent glyphs, we explore simpler hand-copied facsimiles to assess the capabilities of MLLMs. In subsequent research, to enhance practical applicability, we will conduct further studies on ink rubbings. Additionally, the character recognition performance in this study still falls short of practical application requirements. In the future, we will explore more data augmentation methods to enhance the character recognition capabilities of MLLMs. Furthermore, we encourage researchers to explore more and larger closed-source models to compensate for our limitations, as we are only able to test three closed-source models due to cost constraints. Finally, we also believe that integrating semantics during the process of glyph evolution is crucial, yet the relevant corpus remains scarce.

\section*{Acknowledgements}
This work is supported by the National Natural Science Foundation of China (NSFC): “Research on Understanding Ancient Characters Based on Multi-modal Large Models” (Grant No. 62476111), China Postdoctoral Science Foundation Funded Project (Grant No. 2024M761122), Natural Science Foundation of Jilin Province (General Program, Grant No. 20260102295JC), the “Paleography and Chinese Civiliza tion Inheritance and Development Program” Collaborative
Innovation Platform (No. G3829), and the National Social Science Foundation of China (No. 23VRC033).

\bibliography{custom}

\clearpage
\appendix

\section{More Task Details} \label{app:task}
\begin{figure*}[h]
    \centering
    \includegraphics[width=1\linewidth]{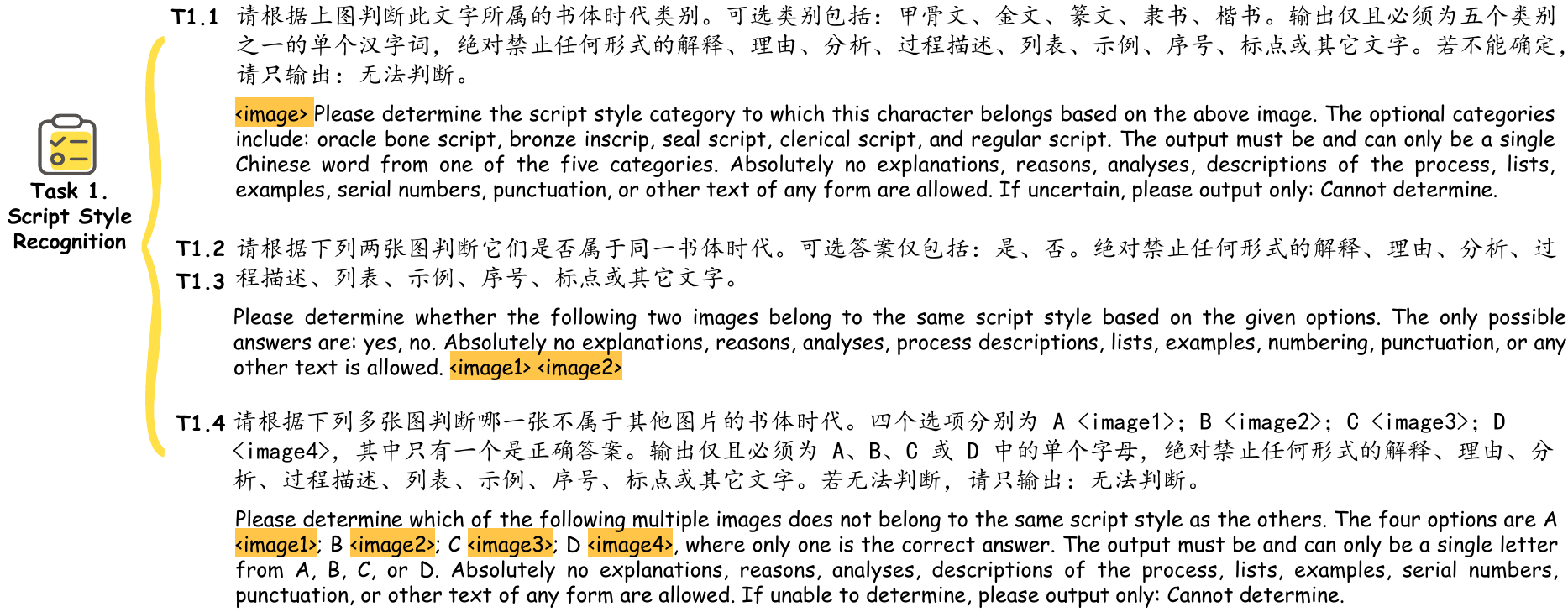}
    \caption{Detailed instructions for Task 1. We use the special character `$<image>$' to represent an image. T1.2 and T1.3 share the same instruction, but in constructing the context for T1.2, it is necessary to ensure that the two images correspond to the same modern Chinese character. In contrast, T1.3 requires ensuring that the images correspond to different modern Chinese characters.}
    \label{fig:tasks1}
\end{figure*}

\begin{figure*}[h]
    \centering
    \includegraphics[width=1\linewidth]{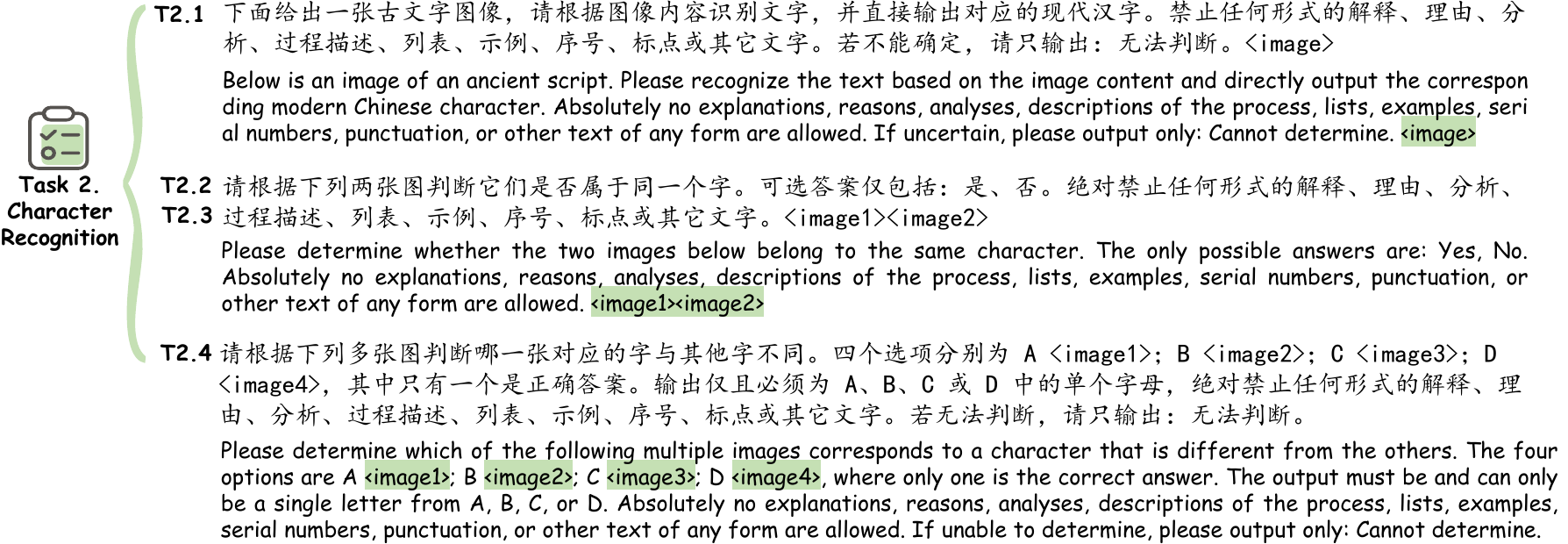}
    \caption{Detailed instructions for Task 2.}
    \label{fig:tasks2}
\end{figure*}

\begin{figure*}[h]
    \centering
    \includegraphics[width=1\linewidth]{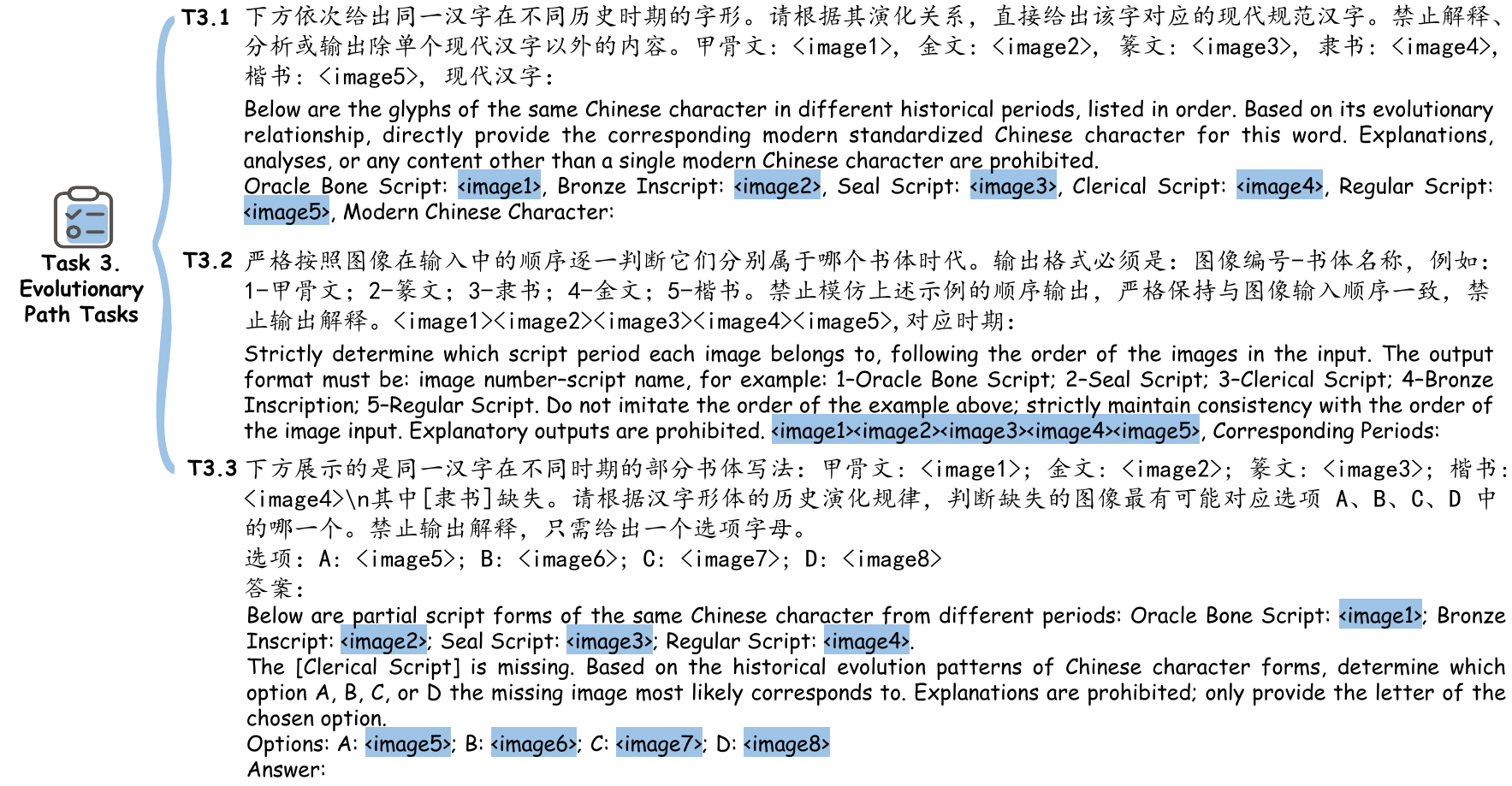}
    \caption{Detailed instructions for Task 3.}
    \label{fig:tasks3}
\end{figure*}
In practical tasks, to standardize the input for the model, we provide longer and more detailed instructions compared to those in Figure~\ref{fig:tasks1}, \ref{fig:tasks2}, and \ref{fig:tasks3}. A more detailed breakdown of the task composition is provided in Figure 3. Each task uses accuracy as the evaluation metric, meaning that if the correct answer appears in the generated result, it is considered correct. If the generated result contains multiple candidate answers, it is assumed that the model did not understand the instruction and is therefore considered a prediction failure. Additionally, for the path comparison task T3.2, given the difficulty of correctly ordering the entire evolutionary path, we adopt a more lenient evaluation strategy: if the prediction for a certain position in the corresponding path is correct, the image is considered correctly predicted. Ultimately, the denominator is the total number of images in the path. For example, if the scrambled evolutionary path of a character is Oracle Bone Script → Regular Script → Bronze Inscription, and the predicted result is "Bronze Inscription → Regular Script → Oracle Bone Script," then the prediction accuracy is 1/3. 

Furthermore, the 11 tasks correspond to different capabilities of the model, and we further explain why these instructions are included in the benchmark.

\begin{itemize}
    \item \textbf{T1.1}. It demonstrates the model's basic glyph recognition capability by determining the script period corresponding to a given character image, which is one of the most fundamental tasks in evolutionary analysis. Therefore, we emphasize placing it at the forefront.

    \item \textbf{T1.2}. It serves as an effective extension of T1, aimed at comparing script styles when presented with different glyphs of the same character. Based on experimental results, providing the character information in advance improves the model's comparative performance.

    \item \textbf{T1.3}. It is also an extension of the basic glyph comparison task, testing the model's discriminative ability by comparing glyphs under the prior condition of different characters. In most cases, it performs better than T1.1 but slightly worse than T1.2. This indicates that the model is more adept at comparing different glyphs of the same character.

    \item \textbf{T1.4}. It is a higher-order extension of T1.1, used for comparison among multiple different glyphs.

    \item \textbf{T2.1}. It represents another fundamental task in ancient script studies: character recognition. Due to the significant changes in glyph forms during the evolution of ancient scripts, the character recognition capability of the model is generally poor.

    \item \textbf{T2.2}. It is an extension of the basic character recognition task, determining whether two glyphs represent the same character within the same script style. 

    \item \textbf{T2.3}. Similarly, it is an extension of the basic capability of T2.1, determining whether two glyphs represent the same character under the premise of different script styles.

    \item \textbf{T2.4}. It is a high-order extension of T2.1, designed to evaluate a model's capability in comparative recognition and selection among multiple characters.

    \item \textbf{T3.1}. This is a character recognition task within an evolutionary path, which encourages the model to utilize more evolutionary context to determine the exact character. Compared to T2.1, this task is simpler because it provides more background knowledge for the model to reference.

    \item \textbf{T3.2}. This is a task of ordering evolutionary sequences, designed to encourage the model to reconstruct the chronological order of glyph evolution within a shuffled evolutionary path. It requires the model to possess basic capabilities in calligraphy style recognition and the ability to reorganize the sequences according to temporal progression. Therefore, it is a more complex form of T1.1.

    \item \textbf{T3.3}. This is also an important task in evolutionary analysis, involving the completion of a path when a specific segment is missing. In this process, the model needs to comprehensively consider factors such as the glyph structure and semantics of the characters, and frame the completion task as a retrieval problem. However, due to the vast search space of glyph retrieval, which is not well-suited to MLLMs, we have reformulated it as a multiple-choice question. 
\end{itemize}

\section{More MLLMs Details} \label{app:mllm}
We evaluate several common MLLMs including: \texttt{TongGu-VL-2B-Instruct}~\cite{cao2025tonggu} (An expert model trained on a cultural heritage dataset, which has been reported to exhibit stronger comprehension of ancient Chinese scripts compared to other models.), \texttt{Qwen2.5-VL} series~\cite{qwen2.5-VL} and \texttt{Qwen3-VL} series~\cite{qwen3technicalreport}, \texttt{InternVL-3\_5} series~\cite{wang2025internvl3_5}, \texttt{MiniCPM-V-2\_6} and \texttt{MiniCPM-V-4\_5}~\cite{yao2024minicpm}, \texttt{GLM-4.1V-9B-Thinking}~\cite{glm2024chatglm}, \texttt{DeepSeekOCR}~\cite{wei2025deepseek}, \texttt{LLaVA-1.5} series~\cite{liu2023improved}. Additionally, we compared three closed-source models: \texttt{GPT-4o-mini}, \texttt{GPT-5-mini}~\cite{hurst2024gpt}, and \texttt{Gemini-3-Flash}~\cite{comanici2025gemini}. All API calls were made through third-party interfaces\footnote{https://api.xi-ai.cn/}. We do not keep detailed statistics on the expenses, but including the cost of model debugging, evaluating the benchmarks on the three models exceeded 500\$. Therefore, we do not explore more closed-source models due to cost constraints. 

\section{More Evaluation Results} \label{app:eval}
\begin{table*}[h]
  \centering
  \small
  \setlength\tabcolsep{3.5pt}
  \renewcommand{\arraystretch}{0.5}
    \begin{tabular}{c|cccc|cccc|ccc|c}  \toprule
    MLLMs& T1.1  & T1.2  & T1.3  & T1.4  & T2.1  & T2.2  & T2.3  & T2.4  & T3.1  & T3.2  & T3.3  & Average \\ \midrule
    \texttt{Qwen3-VL-2B-Instruct} & 17.66  & 58.87  & 53.59  & 25.53  & 20.82  & 70.47  & 73.23  & 25.82  & 30.36  & 17.51  & 45.88  & 39.98  \\
    \texttt{Qwen3-VL-8B-Instruct} & 47.79 & 76.63  & 70.22  & 58.45  & 29.76  & 74.71  & 82.77  & 69.67  & 50.96  & 49.20  & 71.75  & 61.99  \\
    \texttt{Qwen3-VL-30B-A3B-Instruct} & 31.47  & 68.20  & 56.31  & 46.88  & 29.56  & 72.34  & 75.66  & 65.72  & 53.60  & 34.70  & 66.15  & 54.60  \\
    \texttt{InternVL3\_5-1B-HF} & 6.66  & 46.20  & 44.50  & 20.42  & 15.60  & 46.92  & 39.14  & 13.40  & 21.20  & 6.18  & 30.65  & 26.44  \\
    \texttt{InternVL3\_5-8B-HF} & 17.81  & 54.02  & 52.41  & 26.12  & 6.47  & 59.21  & 39.22  & 38.78  & 16.30  & 20.01  & 38.57  & 33.54  \\
    \texttt{MiniCPM-V-2\_6} & 7.87  & 52.31  & 52.61  & 27.71  & 8.40  & 77.61  & 78.02  & 42.19  & 8.06  & 16.61  & 24.80  & 36.02  \\
    \texttt{MiniCPM-V-4\_5} & 21.26  & 63.80  & 54.94  & 35.48  & 9.04  & 80.71  & 78.61  & 62.04  & 15.14  & 25.68  & 24.96  & 24.96  \\
    \texttt{GLM-4.1V-9B-Thinking} & 25.16  & 51.37  & 51.33  & 30.20  & 19.16  & 73.39  & 71.02  & 46.34  & 39.75  & 30.86  & 50.53  & 44.46  \\
    \texttt{DeepSeekOCR-3B} & 10.43  & 51.55  & 49.06  & 24.54  & 8.82  & 52.65  & 55.21  & 20.06  & 19.47  & 14.96  & 46.19  & 32.09  \\
    \texttt{LLaVA-1.5-7B-HF} & 1.58  & 8.31  & 9.23  & 24.94  & 0.07  & 27.49  & 24.89  & 26.76  & 0.11  & 12.18  & 27.51  & 14.82  \\
    \texttt{LLaVA-1.5-13B-HF} & 2.84  & 45.21  & 46.84  & 6.93  & 0.16  & 48.78  & 54.65  & 5.82  & 0.06  & 6.26  & 13.10  & 20.97  \\ \bottomrule
    \end{tabular}%
    \caption{Performance comparison of different MLLMs across all tasks (accuracy \%). Due to cost constraints, we do not evaluate the full dataset on additional closed-source models.} 
  \label{tab:eval2}%
\end{table*}%
Table~\ref{tab:eval2} presents the performance of different MLLMs on reasoning tasks across the entire dataset (including both the training and test sets). Consistent with the results in Table~\ref{tab:eval}, the \texttt{Qwen3} series models achieve the best performance, with \texttt{Qwen3-VL-8B-Instruct} obtaining the highest average performance. Additionally, MLLMs show certain potential for glyph discrimination tasks, though text recognition remains a significant challenge. The consistent trends indicate that the test set distribution aligns with the overall dataset distribution, allowing it to serve as a representative proxy for evaluating model performance on the entire dataset—at only 1/10 of the inference cost.

\begin{figure}
    \centering
    \includegraphics[width=1\linewidth]{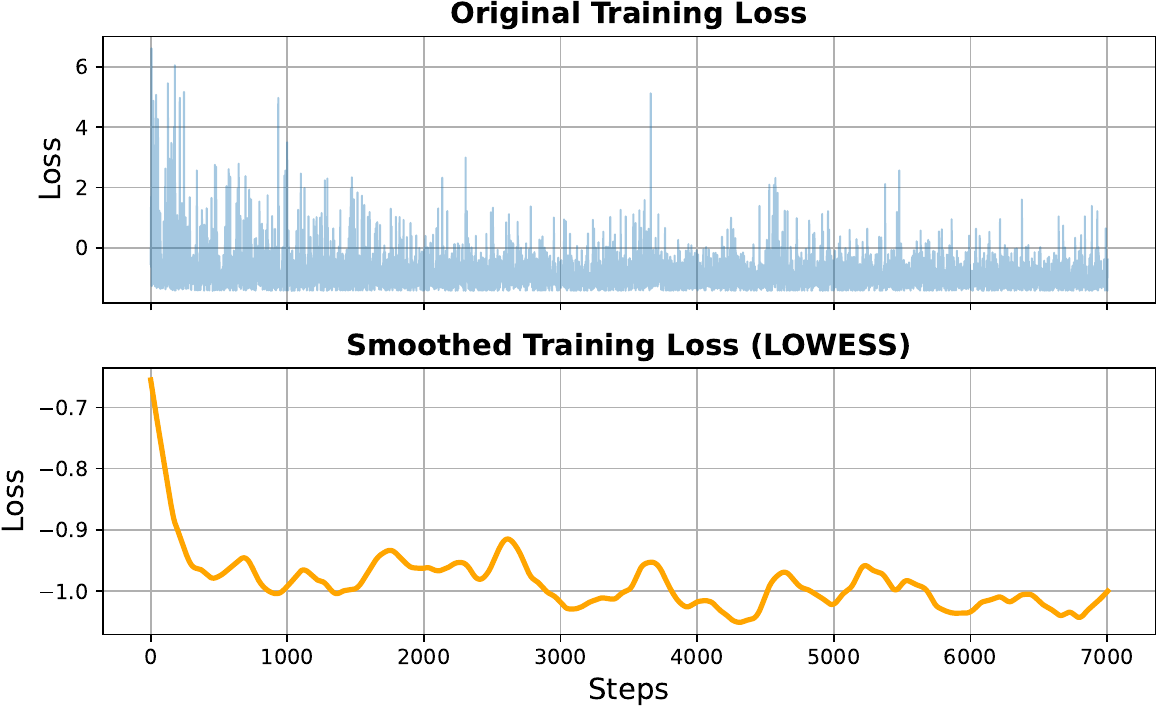}
    \caption{During the first stage, the variation of the contrastive learning loss driven by glyphs and the corresponding Locally Weighted Regression fitting curve.}
    \label{fig:cl_loss}
\end{figure}

\begin{figure}
    \centering
    \includegraphics[width=1\linewidth]{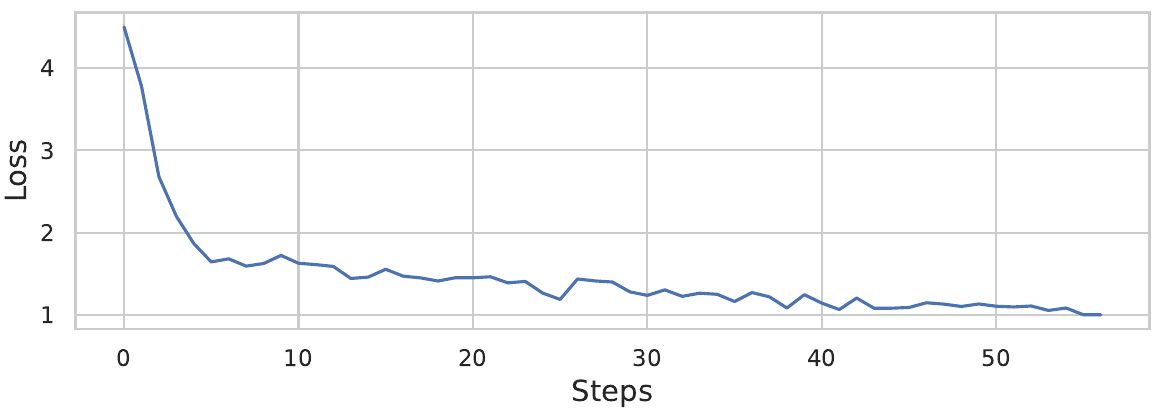}
    \caption{The loss variation of the model in the second stage.}
    \label{fig:loss2}
\end{figure}

\begin{figure}
    \centering
    \includegraphics[width=1\linewidth]{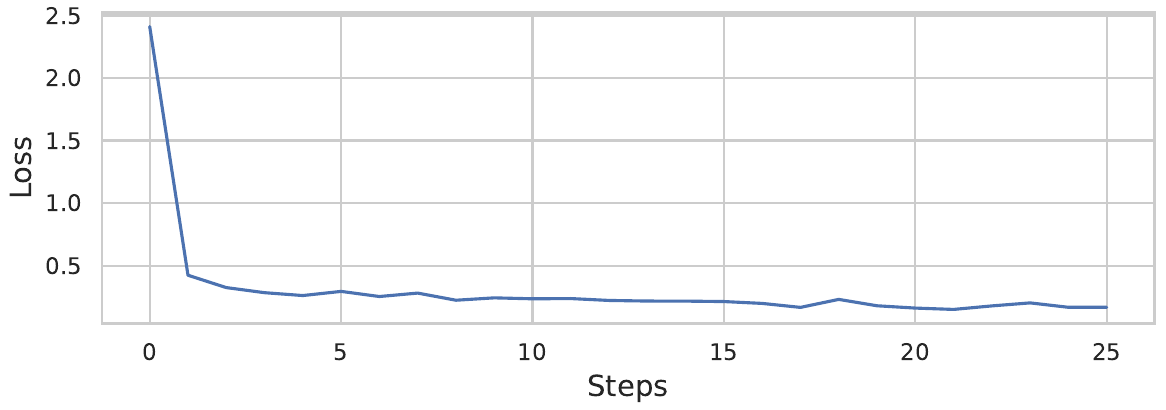}
    \caption{The loss variation of the model in the third stage.}
    \label{fig:loss3}
\end{figure}

\section{Model Fine-tuning Details} \label{app:sft}
We use LlamaFactory\footnote{https://www.llamafactory.cn/} for fine-tuning. During the first stage of fine-tuning, we package each image into a standard context template, which includes two key inputs: $\{``type": ``text", ``text": char \}$ and $\{``type": ``image", ``image": img\}$. Here `char' represents the standard modern Chinese character text, and `img' represents the path to the corresponding image. Subsequently, the language model parameters of MLLMs (taking Qwen3-VL as an example) are frozen, and corresponding image representations are obtained for computing the loss $\mathcal{L}_{con}$. In the second stage, we encapsulate the instructions and answers from T2.1 (drawn exclusively from the training set to prevent information leakage) into a standard training template, and fine-tune the language model component of the MLLMs to obtain the second fine-tuned version. Finally, in the third stage, we fine-tune the language model component across all tasks in the benchmark to obtain the final evolutionary understanding model. 

Figure~\ref{fig:cl_loss} illustrates the variation of the overall loss function during training in the first stage. Although the learning loss exhibits significant fluctuations across different images, the overall fitting curve indicates that the loss function is effectively decreasing.

Figure~\ref{fig:loss2} shows the variation in loss values during the second stage of model training. Compared to the loss from the glyph comparison learning in the first stage, the loss in the second stage is smoother. This is because developing the model's character recognition capability is relatively simpler than comparing glyphs, especially considering the strong reasoning abilities already present in the existing MLLMs. Similarly, Figure ~\ref{fig:loss3} illustrates the loss during the third stage of training. Given the relatively small training dataset and the rapid convergence observed, we argue that SFT data is not required in large quantities for evolutionary analysis in MLLMs.

In the task-specific SFT stage of the second/third phase, we employ the same learning rate of 1e-5 for 3 epochs, with a warmup ratio of 0.1. It should be noted that \texttt{Qwen3-VL-2B-SFT} also follows the same training strategy on identical samples. This process do not employ the LoRA strategy and involve fine-tuning on the full set of parameters. All experiments are conducted on 4*A100 80GB GPUs. We save the model parameters after training and evaluate them across various tasks.

\section{More Visualization Results} \label{app:vis}
\begin{figure}[h]
    \centering
    \includegraphics[width=1\linewidth]{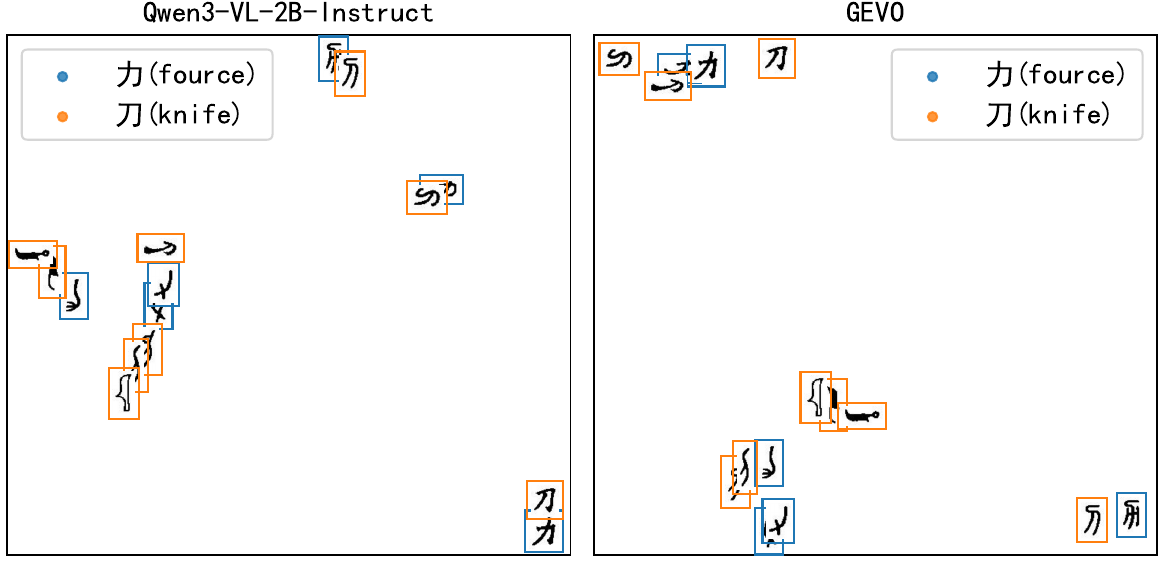}
    \caption{Visualization of similar image representations among ``\begin{CJK*}{UTF8}{gbsn}力\end{CJK*}'' (fource) and ``\begin{CJK*}{UTF8}{gbsn}刀\end{CJK*}'' (knife) baed on \texttt{Qwen3-VL-2B-Instruct} and GEVO in two-dimensional space.}
    \label{fig:vis_li}
\end{figure}

\begin{figure}[h]
    \centering
    \includegraphics[width=1\linewidth]{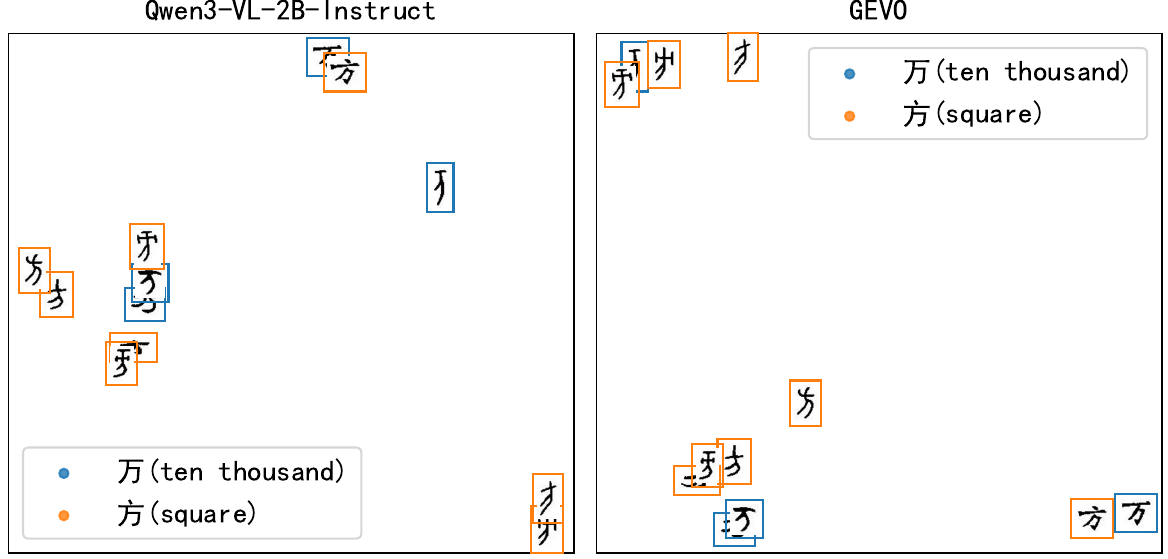}
    \caption{Visualization of similar image representations among ``\begin{CJK*}{UTF8}{gbsn}万\end{CJK*}'' (ten thousand) and ``\begin{CJK*}{UTF8}{gbsn}方\end{CJK*}'' (square) baed on \texttt{Qwen3-VL-2B-Instruct} and GEVO in two-dimensional space.}
    \label{fig:vis_wan}
\end{figure}

Figure~\ref{fig:vis_li} and Figure~\ref{fig:vis_wan} also confirm the influence of glyph similarity on the model. For ``\begin{CJK*}{UTF8}{gbsn}力\end{CJK*}'' (fource) and ``\begin{CJK*}{UTF8}{gbsn}刀\end{CJK*}'' (knife), \texttt{Qwen3-VL-2B-Instruct} exhibits a deficiency in distinguishing characters that share similar glyphs but are actually different characters (\glyph{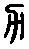} and \glyph{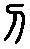}). Correspondingly, GEVO tends to assign larger relative distances to the two. As previously discussed, GEVO still lacks the corresponding capability to distinguish between extremely similar glyphs (\glyph{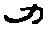} and \glyph{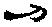}), which is also one of the future research directions. The same phenomenon is also observed between ``\begin{CJK*}{UTF8}{gbsn}万\end{CJK*}'' (ten thousand) and ``\begin{CJK*}{UTF8}{gbsn}方\end{CJK*}'' (square), where GEVO can distinguish the two characters in the bottom right corner and maintain a greater distance, whereas in the results of \texttt{Qwen3-VL-2B-Instruct}, the two characters partially overlap. 

Another interesting finding is that although these characters share similar writing styles during certain evolutionary stages, they possess fundamentally distinct semantics. For example, ``\begin{CJK*}{UTF8}{gbsn}万\end{CJK*}'' (ten thousand) and ``\begin{CJK*}{UTF8}{gbsn}方\end{CJK*}'' (square) differ only slightly in their written composition, but the meanings they convey are vastly different. Based on this, we also hope that our research can inspire professional paleographers to study and explain the aforementioned phenomena from the perspective of MLLMs' understanding of glyphs.

\end{document}